\pdfoutput=1

\documentclass[11pt]{article}

\usepackage{acl}
\usepackage{times}
\usepackage{latexsym}
\usepackage[T1]{fontenc}
\usepackage[utf8]{inputenc}
\usepackage{microtype}
\usepackage{inconsolata}
\usepackage{graphicx}
\usepackage{amsmath}
\usepackage{booktabs}

\hypersetup{
  pdftitle={Mind the Cap: Output-Budget Regimes Change the Measured Multilingual Reasoning Gap},
  pdfauthor={Ankit Goyal and Jaideep Ray}
}

\title{Mind the Cap: Output-Budget Regimes Change the Measured Multilingual Reasoning Gap}

\author{
  Ankit Goyal \\
  \texttt{ankit@goyalankit.com}
  \And
  Jaideep Ray \\
  \texttt{jaray@acm.org}
}

\begin{document}
\maketitle

\begin{abstract}
Multilingual evaluations report accuracy at a single output-token cap, but languages need different numbers of tokens to express the same content. The cap is therefore a hidden experimental variable. We test whether the native-vs-translate gap on MGSM (German, Thai, Swahili) is a token-budget artifact for Qwen3-8B and Llama-3.1-8B-Instruct under four prompting strategies. The measured gap swings by up to 57 points across budgets, length normalization moves it by up to 38.9 points where the cap binds, and at tight caps normalization can reverse which strategy scores higher. We prospectively froze the sweep's three Qwen peaks and its near-zero value at 1024 and evaluated them on 540,000 independently hard-capped decodes. A second frozen family of six Holm-corrected tests rejects every null. The prospectively frozen test at \(B^*=1024\) still fails to reject because native accuracy has already saturated there. Above saturation, the residual difference is a strategy-performance gap, not an identified reasoning deficit. The same truncation channel prices a cost-ordered adaptation ladder: a cross-fitted Thai vocabulary extension closes 0.00 points of the gap at the frozen budget and 4.9 points where 19\% of traces still truncate. A third frozen family varies only the \emph{announced} budget at a fixed enforced cap. Announcing 128 rather than 2048 tokens moves Thai native accuracy by 5.1 points, so accuracy is not a function of the enforced cap alone. Answer-emission timing accounts for where the artifact sits: with \(G(t)\) the probability that a trace is correct and has emitted its answer by \(t\), the estimand identity gives \(\Delta_L(B)=G(\lfloor rB\rfloor)-G(B)\) from one long-cap run. This matches the three pre-specified MGSM peaks to 0.65 points and, on three further benchmarks, tracks held-out items to 0.92 points while locating the peak exactly in five of seven Qwen cells. Treat the output cap as an independent variable and report accuracy across the budget regime, not at a single budget.
\end{abstract}

\section{Introduction}

A recurring finding in multilingual LLM evaluation is that models solve reasoning problems better when the problem is first translated into English than when they reason in the original language. This ``multilingual reasoning gap'' is usually measured at one generation length. Languages differ in how many tokens they need to express parallel content, and strategies differ in when they emit their answer. Under a hard output cap, both features interact with the cap in ways that can resemble a strategy-performance difference.

We ask a narrow, falsifiable question: is the native-vs-translate gap on MGSM a token-budget artifact? We operationalize a budget artifact as the change in that gap between a token-matched cap and a length-normalized, FLORES-200-premium-adjusted cap. If giving NATIVE its premium-scaled budget closes the gap, token matching contributed to the measured difference; if the gap persists, this particular correction does not explain it.

Our contribution is methodological:
\begin{enumerate}
    \item The measured contrast can depend strongly on the evaluation budget. We pair a prospectively frozen non-rejection at one budget with a retrospective sweep that locates the budget-binding regime, then replicate that budget dependence under a pre-specified held-out design in which every budget is decoded under its own hard cap.
    \item Under tight caps, length normalization can reverse which strategy looks better, and announcing a budget changes behavior even when the enforced cap is held fixed.
    \item Five ledger-computable audits---trace-language ID, COMET translation quality, parser robustness, decoder parity, and normalizer sensitivity---bound the interpretations the stored ledger supports.
    \item The same truncation channel prices a cost-ordered adaptation ladder: a directly measured, cross-fitted vocabulary extension closes the gap only where the cap still truncates.
\end{enumerate}

Prompt language, reformulation, answer-format compliance, translation quality, and reasoning-trace language are confounded in this design. The object of inference is therefore strategy performance under controlled prefix budgets, not causal reasoning ability.

\section{Related work}
\label{sec:related}

This study connects multilingual chain-of-thought evaluation on MGSM \citep{mgsm} with work on test-time compute, output budgeting, and budget-forcing, including s1 \citep{s1budget}. Our length proxy follows the FLORES parallel-text lineage and NLLB's FLORES-200 benchmark \citep{flores101,nllb}, while the measurement checks use GlotLID \citep{glotlid} and COMET \citep{comet22}. Cross-lingual tokenization disparities make output length a language-specific computational cost \citep{ahia2023,petrov2023}. A common response is to extend the vocabulary with target-language pieces and continue pretraining \citep{chinesellama,swallow}; principled initialization of the new embeddings is one refinement \citep{focus}, and \S\ref{sec:ladder} prices that intervention against our estimand.

Inference-budget work likewise finds that truncating reasoning harms accuracy in model- and modality-dependent ways \citep{brokenchains}, while training models to recover answers from truncated traces could shrink the budget-binding regime \citep{trsd}. Multilingual studies find that English-pivot reasoning can improve accuracy yet drift off target \citep{mmath}, and identify input comprehension as a major loss source addressed by an English-intermediary pipeline \citep{ust}. Training-side methods further shrink native--pivot gaps to roughly 2--3.5\% under matched supervision \citep{layerswap} and produce cross-lingual self-distillation gains that grow with inference budget \citep{copsd}. Our contribution is orthogonal and evaluational: these methods should be measured under budget sweeps that reveal whether supervision, strategy, or answer-emission timing drives the reported gap.

\section{Design and estimands}
\label{sec:design}

\textbf{Data and strategies.} MGSM has 250 items per language in German (de), Thai (th), and Swahili (sw), three languages that span a wide range of FLORES-200 token premiums. We evaluate four strategies, also called arms. NATIVE reasons and answers in language \(L\). TRANSLATE-ACT translates the problem to English, then solves it. PIVOT reasons in English and answers in \(L\). CODE-SWITCHED uses an English scaffold. We draw \(k=8\) samples per item from Qwen3-8B \citep{qwen3} (confirmatory) and Llama-3.1-8B-Instruct \citep{llama3} (secondary, not a replication), for 48,000 stored generations.

\textbf{Prefix-defined budgets.} Each (item, sample) has one stored 4096-token generation. We call this store of generations, together with the scores of all their prefixes, the \emph{ledger}. Evaluating budget \(B\) means scoring the length-\(B\) prefix. This removes between-budget sampling noise but limits conclusions to prefix-defined evaluation. The retrospective regime sweep uses \(B \in \{64,128,192,256,384,512,768,1024\}\); 2048 and 4096 are added only for extended crossover and best-arm checks.

\textbf{Independent decoding.} Hard-capped decodes separate budget effects from the shared trajectory in prefix scoring. The stored ledger supplies pre-specified point predictions for a fresh confirmation sample; peak budgets are fixed from discovery rather than re-selected. The sample contains 540,000 decodes in 270 cap-partitioned shards, covering both models, all four arms, three languages, and \(k=8\). It evaluates \(B \in \{64,128,192,256,384,512,768,1024,2048\}\) and the premium-scaled NATIVE caps \(\lfloor r_{m,L}B\rfloor\). Seeds vary by cap, no two budgets replay one trajectory, and no trace exceeds its cap.

\textbf{Scoring.} We use strict prefix-only exact match on \texttt{\#\#\#\# <integer>} under intention to treat; truncated, non-integer, and non-compliant answers score 0. Each of the eight samples per item is scored independently at every budget, not by best-of-\(k\), pass@\(k\), or majority vote. Reported accuracy averages all item--sample cells, and the item-clustered bootstrap resamples the 250 MGSM items while retaining all eight samples within each selected item.

\textbf{Estimand.} Let \(\operatorname{acc}_N\) and \(\operatorname{acc}_T\) denote NATIVE and TRANSLATE-ACT accuracy, let \(\operatorname{gap}(B)=\operatorname{acc}_T(B)-\operatorname{acc}_N(B)\), and let \(r_{m,L}\) be the FLORES-200 token premium of language \(L\) over English for model \(m\). Then
\begin{align}
\Delta_L(B)
&= [\operatorname{acc}_T(B)-\operatorname{acc}_N(B)] \notag\\
&\quad - [\operatorname{acc}_T(B)
          -\operatorname{acc}_N(\lfloor r_{m,L}B\rfloor)] \notag\\
&= \operatorname{acc}_N(\lfloor r_{m,L}B\rfloor)
   -\operatorname{acc}_N(B).
\end{align}

TRANSLATE-ACT cancels. What remains is simply how many more NATIVE answers become correct between the matched budget \(B\) and the premium-scaled budget \(\lfloor r_{m,L}B\rfloor\). We call a budget \emph{binding} when correct answer lines are still appearing inside that interval, and the \emph{budget-binding regime} is the range of \(B\) over which they are.

This identity has three direct consequences. First, \(\Delta_L(B)\) is a finite, discrete increment of the NATIVE accuracy curve, not a comparator interaction. Second, its peak location follows the NATIVE answer-emission distribution, while its height depends on the premium-scaled interval and the native curve inside it. Third, \(\Delta_L(B)\to0\) once NATIVE accuracy saturates, so a near-zero value above score saturation follows analytically from the estimand.

\textbf{Primary test.} Each protocol is an internal freeze recorded as a git tag before the data it governs existed, not a public preregistration; no registry filing was made. The frozen evaluation budget is \(B^*=1024\), the largest \(B\in\{512,1024\}\) with every \(\lfloor r_{m,L}B\rfloor\le4096\). The confirmatory family contains six Holm-corrected Qwen tests: H1-existence, directional H1-SESOI (a smallest-effect-size-of-interest test, \(\Delta_L(B^*)>5\) points for at least one language), H2 (a larger effect for Thai than German), and H3 (a strategy reversal on the matched-dollar grid, which equalizes notional serving cost rather than output tokens) for each of the three languages. Inference uses an item-clustered paired bootstrap (10,000 resamples), a studentized sup-\(t\) maximum statistic, and a pre-specified \(1.3\times\) tail-conservatism factor; Appendix~\ref{app:stats} gives details.

FLORES-200 premiums are Qwen de 1.56 / th 2.55 / sw 1.94 and Llama de 1.58 / th 2.19 / sw 1.93. For each model and language, \(r_{m,L}\) is the total-token ratio to parallel English over all 1,012 FLORES-200 devtest sentence pairs after NFC normalization, computed with that model's own tokenizer and a paired percentile bootstrap over sentence pairs.

\section{Regime-dependent results}
\label{sec:results}

\subsection{The frozen test yields no confirmatory support}
\label{subsec:frozen}

At \(B^*=1024\), \(\Delta_L(B^*)\) is near zero for every Qwen language, and all six Holm-family tests fail to reject (formal outcome \texttt{no\_confirmatory\_h1\_support}):
\begin{itemize}
    \item Qwen \(\Delta_L(B^*)\): de 0.00, th 0.15, sw 0.05 points. H1-existence raw \(p=0.060\) at Holm-local \(\alpha=0.0083\); H1-SESOI raw \(p=1.0\).
    \item Llama \(\Delta_L(B^*)\): de/th/sw all 0.00. This procedurally matched secondary analysis is outside the confirmatory family and also rejects nothing.
\end{itemize}

The two non-rejections have different descriptive causes. Qwen NATIVE accuracy plateaus at 79.0\% / 47.1\% / 33.7\% for de/th/sw; between 1024 and the Thai FLORES-scaled budget of 2611, only 3 of 2000 traces become newly correct. Llama NATIVE instead plateaus near floor at 13.6\% / 3.9\% / 29.0\%, alongside never-emission rates of 80.2\% / 93.2\% / 46.0\%. Qwen therefore reaches score saturation at moderate accuracy, whereas Llama de/th often never produces a parseable native answer. Neither pattern implies that all traces have terminated.

\subsection{Tight budgets expose a large, then vanishing, artifact}
\label{subsec:tight}

Table~\ref{tab:peak-accuracy} places the largest retrospective \(\Delta_L(B)\) inside the budget-binding regime (pointwise item-clustered bootstrap 95\% CIs).

\begin{table*}[t]
\centering
\small
\caption{Peak length-normalization effects, where \(\Delta_L(B)=\operatorname{acc}_N(\lfloor rB\rfloor)-\operatorname{acc}_N(B)\), together with values at the frozen budgets. Intervals are pointwise item-clustered bootstrap 95\% CIs. The independent column is the same estimand on the confirmation sample of \S\ref{sec:design}, evaluated at the discovery peak budget rather than at a re-selected argmax.}
\label{tab:peak-accuracy}
\resizebox{\textwidth}{!}{%
\begin{tabular}{lrrrrrrr}
\toprule
model/lang & replay peak \(\Delta_L(B)\) (pts) & independent \(\Delta_L(B)\) & peak \(B\) & NATIVE acc. at \(B\) & TRANSLATE-ACT acc. at \(B\) & \(\Delta_L(512)\) & \(\Delta_L(1024)\) \\
\midrule
Qwen de  & 34.2 [30.2, 38.3] & 34.65 & 192 & 16.10 & 22.60 & 2.25 & 0.00 \\
Qwen th  & 38.9 [34.7, 43.0] & 38.60 & 256 & 6.20  & 47.75 & 8.85 & 0.15 \\
Qwen sw  & 15.0 [12.4, 17.7] & 13.70 & 128 & 8.70  & 0.60  & 0.25 & 0.05 \\
Llama de & 8.4 [7.0, 9.9]    & 8.50  & 256 & 3.85  & 43.10 & 0.15 & 0.00 \\
Llama th & 2.3 [1.6, 3.1]    & 2.10  & 192 & 1.10  & 16.90 & 0.15 & 0.00 \\
Llama sw & 18.2 [15.9, 20.6] & 17.65 & 256 & 8.25  & 44.60 & 1.60 & 0.00 \\
\bottomrule
\end{tabular}
}
\end{table*}

The Qwen rows of Table~\ref{tab:peak-accuracy} contain three different kinds of peak. German gives the clearest two-arm comparison: at \(B=192\) both arms are well away from floor, and premium-scaling raises NATIVE by 34.2 points. Thai has the largest peak and the largest premium: at \(B=256\), \(r=2.550777\) gives a premium-scaled budget of 652, so \(\Delta_L(256)\) summarizes the gain across a 396-token interval rather than a contrast at equal output widths. Swahili is weaker evidence, because its peak occurs where TRANSLATE-ACT scores only 0.60\%, so it shows a native-prefix rescue rather than a clean two-arm comparison away from floor. The same caution applies to the German crossover at \(B=128\) (\S\ref{subsec:crossover}).

Simultaneous max-\(|t|\) 95\% bands over the sweep keep every Qwen peak away from zero (de@192 [27.8, 40.6], th@256 [32.3, 45.4], sw@128 [10.7, 19.3]), and peak locations are stable in 89.6\%, 100.0\%, and 87.9\% of bootstrap replicates. At \(B^*\) the largest language-specific upper bound is 0.32 points for Qwen and 0.00 for Llama, below the exploratory \(\pm5\)-point equivalence margin.

\textbf{The peaks survive independent capping.} The confirmation sample has its own frozen family of six Holm-corrected Qwen tests: three one-sided SESOI tests that \(\Delta_L(B)>5\) points at the discovery peak budget, and three two-one-sided equivalence tests that \(|\Delta_L(1024)|<5\). All six reject at family-wise \(\alpha=0.05\) (formal outcome \texttt{confirmatory\_support}). The independent peak estimates are 34.65 points for German at \(B=192\), 38.60 for Thai at 256, and 13.70 for Swahili at 128. Their standard errors are 2.10, 2.26, and 1.37, with \(p=0.0001\) throughout; each estimate lies inside its discovery interval. At \(B^*=1024\), the equivalence estimates are 0.15, \(-0.25\), and \(-1.25\) points, and the largest of their three p-values is 0.0003. The independent argmax also falls on the predicted budget in all three Qwen languages. This second frozen family does not alter the non-rejection of the \(B^*=1024\) family in \S\ref{subsec:frozen}. Instead, it rules out shared prefix trajectories as the source of the tight-budget magnitudes.

Llama remains outside the confirmatory family and supports no confirmatory claim. Its independent estimates at the discovery peaks are 8.50 points for German, 2.10 for Thai, and 17.65 for Swahili, against discovery values of 8.35, 2.30, and 18.20. The Thai SESOI test does not reject: a discovery estimate of 2.30 points cannot clear a 5-point SESOI. Llama German and Thai each shift their argmax by one grid point, and those cells are too flat to fix a location; Thai reads 2.20, 2.30, and 2.00 at 128, 192, and 256. Per-cell mean output lengths of the independent decodes match the truncated discovery traces to a median absolute difference of 0.11\% of the cap for Qwen and 0.16\% for Llama.

\textbf{Normalizer sensitivity.} The behavioral trace-length ratio is smaller than the FLORES premium in all six model-language cells, with non-overlapping pointwise intervals, so FLORES consistently grants the larger premium. Substituting the behavioral ratio would still produce a 5-point artifact in four of the six cells; Qwen Swahili is the exception, and Llama Thai never reaches 5 points even at \(1.5\times\) FLORES. These thresholds are grid-selected and conditional on the stored traces, and the behavioral ratio is not a validated normalizer, because it compares traces that differ in content, correctness, and stopping behavior (Appendix~\ref{app:ratio}).

\begin{figure*}[t]
\centering
\includegraphics[width=\textwidth]{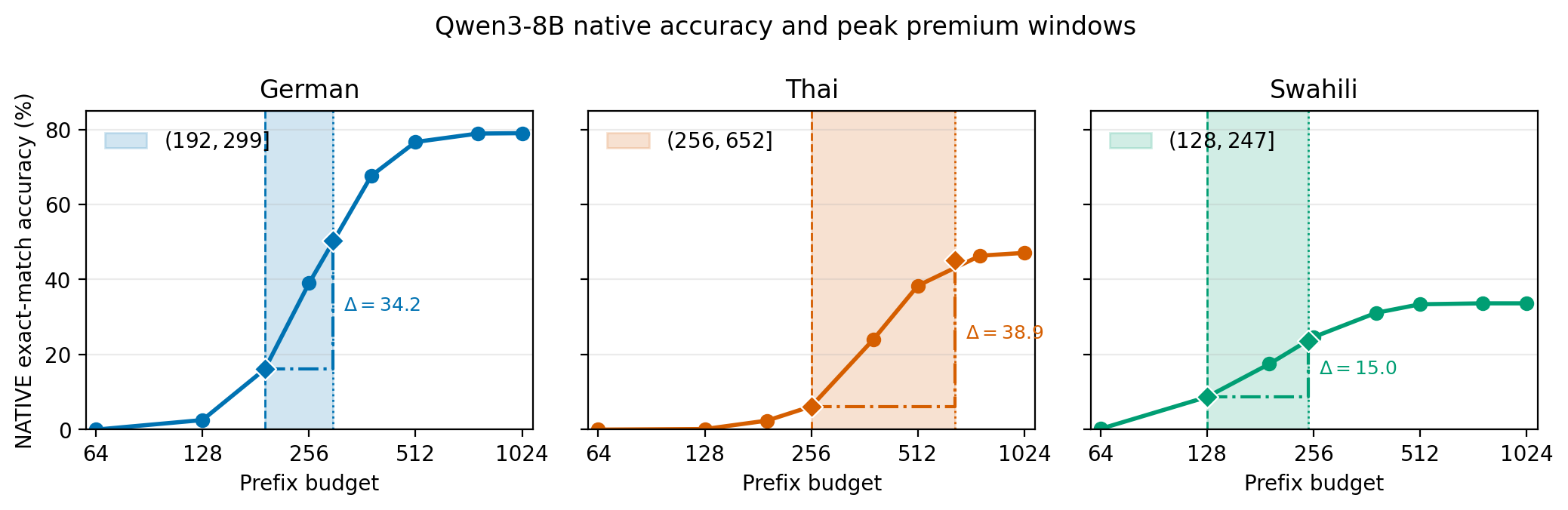}
\caption{Qwen NATIVE accuracy by budget. Shaded intervals mark each language's peak normalization interval \((B_{\mathrm{peak}},\lfloor rB_{\mathrm{peak}}\rfloor]\); the vertical gain between the endpoint markers equals the peak \(\Delta_L(B)\).}
\label{fig:native-curves}
\end{figure*}

\begin{figure*}[t]
\centering
\includegraphics[width=\textwidth]{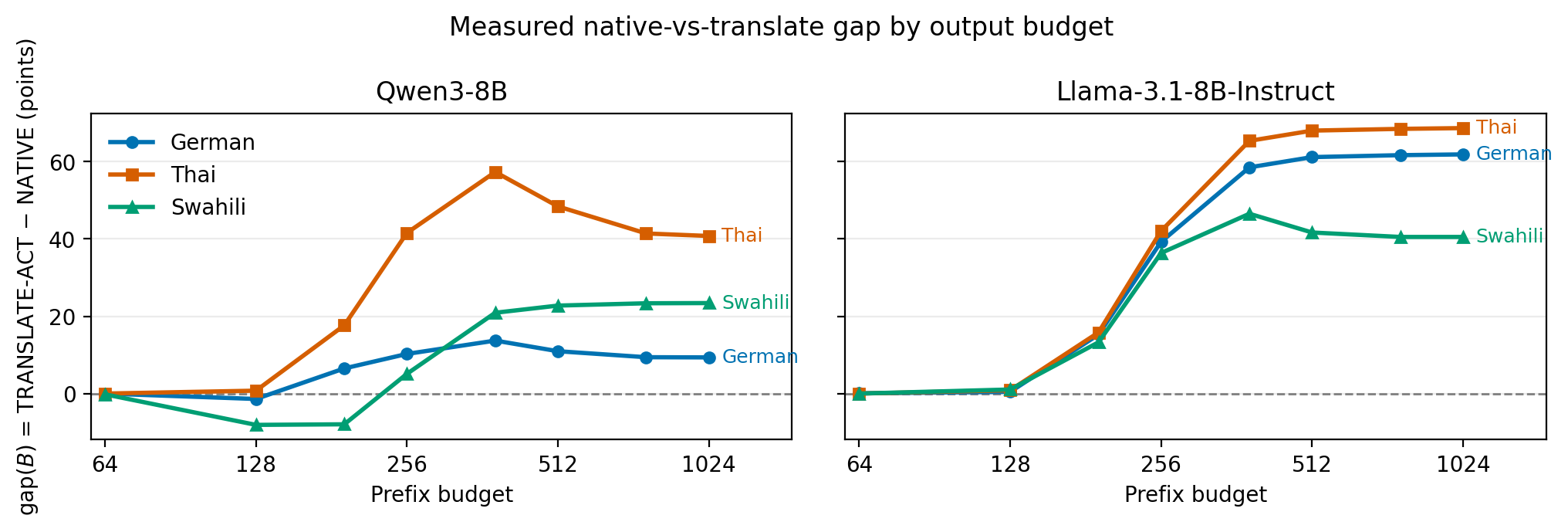}
\caption{The measured gap \(\operatorname{gap}(B)=\operatorname{acc}_T(B)-\operatorname{acc}_N(B)\) by output budget, token-frame prefix accuracies from the stored ledger. Negative values at tight caps are the crossovers of \S\ref{subsec:crossover}; the plateaus at large \(B\) are the saturated strategy-performance gaps of \S\ref{subsec:frozen}. The Qwen Thai gap peaks near \(B=384\), where TRANSLATE-ACT has largely saturated while NATIVE is still emitting answers.}
\label{fig:gap-curves}
\end{figure*}

Figure~\ref{fig:native-curves} plots the Qwen NATIVE curves and marks both accuracy endpoints of each peak premium interval. Figure~\ref{fig:gap-curves} plots the unnormalized gap itself for both models: negative where NATIVE leads under tight caps (\S\ref{subsec:crossover}), steepest while answer emission binds, and flat once accuracy saturates (\S\ref{subsec:frozen}).

\subsection{Tight caps can reverse strategy rankings}
\label{subsec:crossover}

Table~\ref{tab:emission-timing} shows that Qwen crossover strength follows the left tail of answer-emission timing more closely than the medians.

\begin{table*}[t]
\centering
\small
\caption{Qwen answer-emission timing and observed strategy crossovers. \(E\) is the token position at which a parseable answer is emitted; p10 is its 10th percentile.}
\label{tab:emission-timing}
\resizebox{\textwidth}{!}{%
\begin{tabular}{lrrrrl}
\toprule
lang & NATIVE median \(E\) & TRANSLATE-ACT median \(E\) & NATIVE p10 \(E\) & TRANSLATE-ACT p10 \(E\) & observed crossover \\
\midrule
sw & 206 & 262 & 96  & 170.7 & strongest: NATIVE leads at 64/128/192 \\
de & 270 & 247 & 160 & 158   & marginal: NATIVE leads at 128 \\
th & 377 & 250 & 236 & 165   & none \\
\bottomrule
\end{tabular}
}
\end{table*}

The p10 ordering is timing-consistent with the crossover: Swahili has a 74.7-token NATIVE early-tail advantage and the strongest native lead, German differs by only 2 tokens and has a narrow lead, and Thai NATIVE is 71 tokens later and never leads. These grid-resolved summaries do not isolate translation-segment length or establish mediation.

At \(B=128\), Qwen German NATIVE scores 2.55\% versus 1.15\% for TRANSLATE-ACT; Swahili NATIVE leads with bootstrap probability 1.00 at 128 and 192 (transition [192,256]), while German leads with probability 0.958 at 128 (transition [128,192]). Thai has no native lead, and no Llama language crosses because native accuracy remains near floor. The Qwen Swahili crossover also comes from a degenerate heavy-tail cell: NATIVE output-length p90 is 4096 and 25.1\% never emit a parseable answer. The frozen H3 test did not detect these reversals because it examined only the looser pre-specified budgets.

A sharper consistency check uses the correct-emission sub-CDF \(G(t)=P(C=1,E\le t)\), where \(C\) is completed-trace correctness. Equation~(1) gives \(\Delta_L(B)=G(\lfloor rB\rfloor)-G(B)\) from one long-cap NATIVE ledger. This is a consistency check, not a test. For the three Qwen MGSM peak cells, predictions against independent-decoding outcomes have mean absolute error 0.65 points (Appendix~\ref{app:subcdf}). Across three added benchmarks, an exploratory Qwen-only, item-level split-half analysis in the replay frame has MAE 0.92 points on held-out items and locates the peak exactly in five of seven cells (Appendix~\ref{app:subcdf}).

\subsection{Announcing the budget changes behavior}
\label{subsec:announce}

The budget sweeps enforce but do not disclose the cap: \texttt{max\_tokens} stops decoding and does not enter the prompt. A separate frozen family holds the enforced cap at a non-binding 2048 and varies only the \emph{announced} number over \(\{128,256,2048\}\). Its 192,000 fixed-cap records therefore isolate disclosure from truncation. The confirmatory family contains four two-sided announcement dose contrasts on Qwen3-8B: NATIVE and TRANSLATE-ACT crossed with German and Thai. Tests are Holm-corrected at family-wise \(\alpha=0.05\), with first-step local \(\alpha_1=0.0125\) (Appendix~\ref{app:announce}).

One cell rejects, and the formal outcome is \texttt{announcement\_effect\_detected}. Thai NATIVE scores 63.20\% when 128 is announced against 58.10\% when 2048 is, a difference of \(+5.10\) points (SE 1.67, \(p=0.0029\)), and the announced grid is monotone: 63.20, 59.90, 58.10 at 128, 256 and 2048. German NATIVE moves 2.85 points the other way and does not clear its local threshold (\(p=0.0380\)). Announcing a tighter budget therefore \emph{raised} accuracy in the one cell that rejects. That is one cell, one model and one benchmark, and we report it as such rather than as a direction.

Two exploratory results bound the mechanism. A machine-readable \texttt{TOKEN\_BUDGET: \{budget\}} tag is inert in all twelve cells, changing median length by at most 2.8\% in absolute value and rejecting nowhere. Forcing---injecting the answer delimiter when the cap arrives---instead lifts Qwen NATIVE German at \(B=128\) from 2.55\% to 25.70\%, a pooled figure over two populations we keep separate because their average means little: 23.72\% among traces the cap truncated and 100\% among traces that ran to completion without emitting an answer line.

\section{Measurement audits}
\label{sec:audits}

We report five ledger audits and one diagnostic. Trace-language ID, COMET, parser robustness, and decoder parity appear below; normalizer sensitivity appears in \S\ref{subsec:tight}. The verbosity/failure-tail decomposition is a diagnostic, not a sixth audit.

\textbf{Trace-language ID (automated GlotLID).} Automated labels support the main NATIVE vs TRANSLATE-ACT contrast. Determinate NATIVE traces are classified in \(L\) at Qwen de 92.1\% / sw 94.1\% / th 99.4\%, and at 100\% for all Llama languages; TRANSLATE-ACT post-delimiter reasoning is 98.4--99.9\% English.

In the native-Swahili validation cell, a \texttt{swh}-only mapping yields 75\% agreement, below the frozen 90\% criterion. The linguistically correct Swahili macrolanguage mapping (\texttt{swh}+\texttt{swc}) raises Qwen native-Swahili compliance from 85.8\% to 94.1\% and the validation cell to 90.00\% (18/20). Because this mapping was adopted after inspecting the failed criterion, we treat it as a post-hoc analytic decision. An independent blind LLM adjudication of 240 Qwen traces agrees at 96.7\% overall and at least 90\% in every cell, but the frozen human validation remains outstanding. PIVOT and CODE-SWITCHED violate their English instruction in 9 of 12 cells, so they remain outside the main comparison.

\textbf{COMET translation quality.} Reference-based COMET is generally high but nonuniform (Table~\ref{tab:comet}): German is strongest for both models (0.877 Qwen, 0.872 Llama), Qwen Swahili is lowest at 0.749, and Llama Thai has a weak lower tail (p10 0.325). At the six prespecified peak budgets, an exploratory per-cell analysis finds weak and inconsistent COMET--correctness-gain associations (Spearman \(\rho=-0.143\) to \(0.280\), five positive, only one with \(|\rho|\ge0.20\); Table~\ref{tab:comet-gain}), so the tight-budget TRANSLATE-ACT advantage is not well explained by translation quality. Translation quality remains part of the strategy bundle; these scores are descriptive and never condition accuracy.

\textbf{Parser robustness.} In NATIVE peak cells, prefix-only rescued-correct traces are at most 0.35\% and value-unstable traces at most 0.30\%. Within the \((B,\lfloor rB\rfloor]\) intervals, 96.8--100\% of native gains are genuinely terminated. Requiring a terminated answer line changes every peak by at most 0.2 points: Qwen de 34.2 to 34.0, th 38.9 to 38.9, sw 15.0 to 15.1, and Llama sw 18.2 to 18.4. The tight-budget effect is therefore late answer emission rather than a prefix-parser artifact.

\textbf{Decoder parity.} A stratified 2,520-prefix audit finds 37.9\% raw exact-string agreement between local and vLLM decoding but 100\% agreement after production special-token normalization. All raw divergences are cosmetic special-token markup.

\textbf{Verbosity/failure-tail diagnostic.} Qwen Swahili NATIVE has a heavy tail: 10.6\% hit the 4096 cap and 25.1\% never emit a parseable answer. This mixes truncation, non-integer or multiple answers, and format noncompliance, cautioning against interpreting native accuracy as pure reasoning ability.

\textbf{Instrument validity.} A null is interpretable only if the announcement changes behavior. The 30\% median-length-reduction gate excludes both TRANSLATE-ACT cells under instrument v0, at 14.6\% for German and 10.1\% for Thai, so those nulls are uninformative. The translation segment, measured in the pilot, remains at 57 tokens in German and 76 in Thai across announcements. Under instrument v1, total responses reach 34.4\% and 37.5\%, and the estimates are \(-2.60\) (\(p=0.0229\)) and \(-2.30\) (\(p=0.0662\)). No Holm decision or formal outcome changes, but both nulls become interpretable (Appendix~\ref{app:instrument}). Llama fails the gate in all four cells under both instruments, at 2.4--9.3\% against Qwen's 34--43\% under instrument v1; every Llama estimate is therefore uninformative about budget sensitivity, including the two that reject. Llama carries no confirmatory claim.

\section{Implications for adaptation}
\label{sec:ladder}

A cost-ordered adaptation ladder raises the serving budget, changes the
prompting strategy, adds language-specific tokens to the tokenizer, and
finetunes only if all three fail. We price the two token-count rungs.

\textbf{Both token-count rungs act only by relieving truncation.} A larger cap
and a cheaper tokenizer both change one thing: how much of the trace the model
is allowed to finish. Where the trace already fits, neither can change an
answer. This is the same channel Equation~(1) describes: writing \(\rho\) for
the factor by which an extension shortens target-language text, a NATIVE-only
payoff of \(\operatorname{acc}_N(\lfloor\rho B\rfloor)-\operatorname{acc}_N(B)\)
has exactly the form of Equation~(1) with \(\rho\) in place of \(r_{m,L}\). We
use that only as motivation, because compression is not uniform across traces
and a deployed extension changes both arms.

\textbf{Measurement, not extrapolation.} We built genuinely extended Qwen
tokenizers---base vocabulary and merge list untouched, new byte-level merges
learned on target-language traces and appended so that base merges retain
priority---and then measured what they buy under a cap. Each stored trace is
retokenized with the extended tokenizer, its first \(B\) extended token ids are
decoded, and that text is scored with the same strict parser used throughout
the paper. The baseline is the identical computation with the base tokenizer,
and no uniform compression factor is applied. Extensions are cross-fitted over
two disjoint halves of the item set, so no evaluated item contributed a merge,
and the extension size is fixed in advance by a rule that never consults
accuracy. This yields roughly
3.4k, 6.7k and 3.1k new tokens for de/th/sw, which cut the FLORES-200 devtest
premium from 1.559 to 1.531, from 2.551 to 2.207, and from 1.936 to 1.846,
averaged over the two cross-fitting folds (per-fold values in
Table~\ref{tab:vocab}), while the aggregate English devtest token count changes
by less than \(0.02\%\). The extension is applied to both arms, so \(G_3\) reflects one
deployment change rather than a NATIVE-only advantage, and this is a
token-count-only counterfactual rather than a prediction about a retrained
model (Appendix~\ref{app:vocab}).

\begin{table*}[t]
\centering
\small
\caption{Adaptation triage for Qwen. \(G\) is the NATIVE deficit against
TRANSLATE-ACT at the deployed cap \(B\); \(G(4096)\) is the deficit at the
largest stored prefix; \(G_3\) is the deficit after a cross-fitted
language-specific vocabulary extension applied to both arms at the original
cap. ``NAT.\ trunc.'' is the share of NATIVE traces longer than \(B\); ``NAT.\
gain'' is \(\operatorname{acc}_N(4096)-\operatorname{acc}_N(B)\), the NATIVE
accuracy actually recovered by extending the prefix to the largest one stored.
It is observed on this ledger, not a general ceiling: 4096 still binds, and it
does not bound gap closure, which also depends on TRANSLATE-ACT. Gap closure carries item-clustered bootstrap
95\% CIs, conditional on the two fitted tokenizers. The prompting rung is not
priced: the comparator TRANSLATE-ACT \emph{is} that rung, so it closes \(G\) by
construction.}
\label{tab:ladder}
\begin{tabular}{lrrrrrrl}
\toprule
lang & \(B\) & NAT.\ trunc. & NAT.\ gain & \(G\) & \(G(4096)\) & \(G_3\) (vocab) & gap closed by vocab [95\% CI] \\
\midrule
de & 128  & 97.1\% & 76.45 & \(-1.40\)  & \(+9.40\)  & \(-2.30\)  & \(+0.90\) [\(-0.05\), 1.85] \\
de & 256  & 52.1\% & 40.00 & \(+10.25\) & \(+9.40\)  & \(+9.05\)  & \(+1.20\) [\(-0.25\), 2.65] \\
de & 512  & 3.9\%  & 2.35  & \(+10.95\) & \(+9.40\)  & \(+10.45\) & \(+0.50\) [0.05, 1.10] \\
de & 1024 & 0.2\%  & 0.00  & \(+9.35\)  & \(+9.40\)  & \(+9.35\)  & \(+0.00\) [0.00, 0.00] \\
\midrule
th & 128  & 99.9\% & 47.10 & \(+0.75\)  & \(+40.60\) & \(-0.10\)  & \(+0.85\) [0.20, 1.65] \\
th & 256  & 84.7\% & 41.05 & \(+41.55\) & \(+40.60\) & \(+38.15\) & \(+3.40\) [1.90, 5.00] \\
th & 512  & 18.9\% & 8.90  & \(+48.40\) & \(+40.60\) & \(+43.50\) & \(+4.90\) [3.65, 6.20] \\
th & 1024 & 0.7\%  & 0.15  & \(+40.75\) & \(+40.60\) & \(+40.75\) & \(+0.00\) [0.00, 0.00] \\
\midrule
sw & 128  & 81.5\% & 25.05 & \(-8.10\)  & \(+23.30\) & \(-9.40\)  & \(+1.30\) [0.75, 1.90] \\
sw & 256  & 42.6\% & 9.00  & \(+5.00\)  & \(+23.30\) & \(+4.35\)  & \(+0.65\) [\(-0.20\), 1.55] \\
sw & 512  & 15.3\% & 0.35  & \(+22.75\) & \(+23.30\) & \(+22.80\) & \(-0.05\) [\(-0.25\), 0.15] \\
sw & 1024 & 11.6\% & 0.10  & \(+23.40\) & \(+23.30\) & \(+23.35\) & \(+0.05\) [0.00, 0.15] \\
\bottomrule
\end{tabular}
\end{table*}

Table~\ref{tab:ladder} prices the ladder, and three patterns cut against the
expectation that each successive rung shrinks the deficit
(\(G>G_1>G_2>G_3\), writing \(G_k\) for the deficit after applying rung \(k\)).

First, \emph{more budget need not shrink the gap}. At \(B=128\) the measured
deficit is negative in German and Swahili and near zero in Thai, because
TRANSLATE-ACT has not finished its translation preamble
(\S\ref{subsec:crossover}); at the largest stored prefix Swahili moves from
\(G=-8.10\) to \(+23.30\). These cells sit near the accuracy floor, so this is
an exploratory sign instability rather than a calibrated estimate of harm. It
is nonetheless enough to show that a gap measured under a tight cap can carry
the wrong sign.

Second, \emph{vocabulary extension captures only a fraction of what a larger
cap would}. The informative comparison is not the truncated share but
\(\operatorname{acc}_N(4096)-\operatorname{acc}_N(B)\), the NATIVE accuracy a
longer prefix actually recovers: truncated traces that were never going to be
scored correct cannot be rescued by any token-count intervention. Against that
quantity, the extension's own NATIVE gain is 3.45 of 40.00 points for German at
\(B=256\) (9\%) and 5.20 of 41.05 for Thai (13\%), rising to 5.05 of 8.90 for
Thai at \(B=512\) (57\%). Its largest gap closure anywhere in our data is 4.90
points. Where the longer prefix recovers nothing the extension does too: at
\(B=1024\) the NATIVE gain to 4096 is 0.00 / 0.15 / 0.10 points and closure is
\(0.00/0.00/0.05\). The Swahili rows show why the truncated share alone is
misleading: at \(B=1024\), 11.6\% of NATIVE traces exceed the cap, yet almost
nothing is left to recover within the ledger (Appendix~\ref{app:vocab}).

Third, \emph{the residual gap is the whole gap}. At the largest stored prefix,
9.4 / 40.6 / 23.3 points remain for de/th/sw. We cannot price the prompting
rung, because adopting TRANSLATE-ACT closes \(G\) by definition; its standalone
value would have to be established against a different comparator.

The practical guidance is therefore narrow. Before paying for either
token-count rung, measure the realized gain directly: extend the cap on a
sample and see how much accuracy the longer prefixes actually recover. Where
that gain is small, neither rung is likely to pay, however many traces the cap
truncates---though this is dispositive only if the extended cap is itself
non-binding, which ours is not. Where the gain is large, our data do not
support a general ordering. Raising the cap dominates for \emph{native
accuracy}---40.00 against 3.45 points for German at \(B=256\)---but not
necessarily for the \emph{gap}, because a larger cap lifts TRANSLATE-ACT too.
We did not test the finetuning rung, the prompting rung is confounded with our
comparator, and this is one model family on one benchmark, so this is a triage
heuristic derived from our estimand, not a validated adaptation method.

One further scope condition attaches to it. Reading a realized gain off a cap
extension presupposes that \(\operatorname{acc}_N(B)\) is a function of \(B\)
alone. Section~\ref{subsec:announce} shows that it is not, once \(B\) is
announced to the model: one enforced cap can yield different accuracy depending
on what the prompt says about it, so a deployment that discloses its budget
must run the triage under the disclosure it will ship with. This bounds where
the heuristic applies. It does not disturb the truncation argument above, which
quantifies over the cap and the tokenizer, and an announcement is neither.

\section{Scope and implications}
\label{sec:scope}

In these stored-prefix MGSM evaluations, multilingual exact-match comparisons are regime-dependent: length normalization has large descriptive effects while answer emission binds, and negligible effects after score saturation. The prospectively frozen non-rejection at \(B^*=1024\) is a negative boundary condition. The retrospective sweep locates where the sensitivity actually occurs, and its six pre-specified cells then held on independently capped decodes (\S\ref{subsec:tight}). The remainder of the sweep is still exploratory.

At the larger evaluated budgets the residual native-vs-translate difference is large (Qwen Thai about +41 points, Llama Thai about +69), but it is a strategy-performance gap, and prompt language, reformulation, format compliance, translation quality, and reasoning language remain confounded. Calling this residual ``not an identified reasoning deficit'' states what our design can identify, not that execution effects are absent: complementary diagnostics find genuine target-language reasoning-execution failures even with English inputs \citep{datg}.

Future work should test prompt interventions that elicit earlier answer emission---which would shrink NATIVE's budget-binding regime---and extend the protocol beyond MGSM numeric exact match to longer-form and non-numeric tasks.

\textbf{Data and code availability.} We intend to release the analysis code and the per-record score ledgers that reproduce the reported numbers, and the full stored generations as licensing and hosting permit. All benchmarks (MGSM, FLORES-200, Belebele, Global-MMLU-Lite, MMATH) and both models are publicly released artifacts used under their respective licenses; released artifacts will carry attribution and the corresponding license notices.

\textbf{Takeaway.} When comparing language strategies under a token cap, treat the cap as an independent variable rather than a background constant. Report accuracy across a budget sweep with a length-normalized comparison, mark the region in which answer emission binds, and do not let a single budget carry the claim.

\section*{Limitations}

The three Qwen peak cells and the equivalence at \(B^*\) were pre-specified and confirmed on independently capped generations (\S\ref{subsec:tight}). The rest of the sweep remains retrospective and exploratory: the crossovers in \S\ref{subsec:crossover}, grid points outside those six cells, and normalizer-sensitivity analysis. The announcement experiment isolates disclosure at a fixed cap; it does not test an announced budget that also binds, nor establish effects beyond Qwen, MGSM, and the announced values evaluated. The forcing results are exploratory. Under our serving stack \texttt{max\_tokens} only stops decoding and never conditions the model: with a shared seed, 75\% of capped decodes return bitwise identical to the truncated long decode. Separately, repeating an identical request was only 46\% bitwise deterministic (23/50), which weakens reproducibility and shared-seed interpretation without changing the stored-ledger estimand.

The scope is MGSM, three languages, and two 8B models, plus exploratory Qwen-only consistency checks on three added benchmarks. This is not a general claim about multilingual reasoning. The frozen human GlotLID validation and same-content trace-premium validation remain outstanding; the preliminary blind LLM agreement and the behavioral ratio in Appendix~\ref{app:ratio} do not substitute for them. The vocabulary extension is a Qwen-only, token-count counterfactual on fixed emitted text, not a prediction for a retrained model. Its compression is specific to MGSM reasoning traces, its intervals condition on two fixed cross-fitted tokenizers, and \(G(4096)\) is the gap at the largest stored prefix rather than a demonstrated non-binding regime.

\bibliography{custom}

\appendix

\section{Accuracy-curve summary}
\label{app:curves}

The complete curves use \(B\in\{64,128,192,256,384,512,768,1024\}\). Qwen NATIVE accuracy rises from 0.0/0.0/0.2\% at \(B=64\) to 79.0/47.1/33.7\% at \(B=1024\) for de/th/sw; Llama NATIVE rises from 0.0/0.0/0.0\% to 13.6/3.9/29.0\%. The Qwen curves climb sharply through the peak intervals in Figure~\ref{fig:native-curves} and then flatten, whereas the Llama de/th curves remain near floor. The supplementary material reports every model--language--arm value at all eight budgets.

\section{Serving and decoding configuration}
\label{app:compute}

Both models are 8B-parameter instruction-tuned checkpoints served with vLLM
0.17.0 in bfloat16, without quantization, on NVIDIA H100 hardware. Decoding
samples at temperature 0.6 with \(k=8\) samples per item under a 4096-token
generation cap, and Qwen3's thinking mode is disabled
(\texttt{enable\_thinking=false}) so that all reasoning occupies the visible,
budgeted channel. Seeds are item- and sample-specific---the first 64 bits of
SHA-256(base\_seed \(\|\) item\_id \(\|\) sample index)---reused across arms,
languages, and models for pairing and never reused across item--sample pairs;
in the independent-decoding sample they additionally vary by cap
(\S\ref{sec:design}). Truncation is taken from the engine's
\texttt{finish\_reason}, never inferred from the final token id. Decoding
across all experiments totals roughly 466M output tokens, about 22
H100-hours.

\section{Best-English-arm}
\label{app:bestarm}

Reselecting the empirically best English arm inside each bootstrap replicate reproduces the preselected TRANSLATE-ACT gap closely (Qwen Thai about 41 points, Llama Thai about 69; uplift over TRANSLATE-ACT is near zero in most cells).

\section{Trace-length ratio (not a normalizer)}
\label{app:ratio}

The ratio of median NATIVE output tokens to median TRANSLATE-ACT post-delimiter English-reasoning tokens is Qwen de 1.47 / th 2.04 / sw 1.18. Subtracting the FLORES premium gives a negative difference in all six cells, with non-overlapping pointwise intervals: Qwen de \(-0.092\) [\(-0.144\), \(-0.028\)], th \(-0.513\) [\(-0.588\), \(-0.440\)], sw \(-0.757\) [\(-0.824\), \(-0.685\)]; Llama de \(-0.149\) [\(-0.193\), \(-0.099\)], th \(-0.427\) [\(-0.488\), \(-0.365\)], sw \(-0.083\) [\(-0.129\), \(-0.021\)]. This ratio compares behaviorally different traces and cannot validate or replace FLORES.

The minimum evaluated premium that produces a 5-point artifact lies below the behavioral ratio for Qwen de (1.089 vs 1.467) and th (1.188 vs 2.038), and for Llama de (1.274 vs 1.432) and sw (1.253 vs 1.848). Qwen Swahili is the exception: its threshold of 1.254 sits above the behavioral ratio of 1.179, so the 5-point Swahili claim depends on a premium above that ratio, including the frozen FLORES token premium. Llama Thai never reaches 5 points, even at \(1.5\times\) FLORES.

\section{Statistical machinery and the six-test family}
\label{app:stats}

The Qwen confirmatory family contains exactly six raw p-values, corrected by Holm at family level \(\alpha=0.05\). H1-existence tests whether any language has \(\Delta_L>0\); H1-SESOI separately tests whether any has \(\Delta_L>5\) points; and H2 tests the directional ordered contrast \(\Delta_{\mathrm{th}}-\Delta_{\mathrm{de}}>0\). H3-de, H3-th, and H3-sw each test for a dollar-matched strategy reversal, requiring a significantly positive NATIVE-minus-TRANSLATE-ACT contrast at one common-support budget and a significantly negative contrast at another. Each H3 p-value is the maximum of the multiplicity-controlled positive- and negative-side p-values, an intersection-union test.

Inference resamples the 250 parallel MGSM items as clusters, retaining all languages, arms, and eight samples within each selected item, for 10,000 paired bootstrap resamples. Studentized sup-\(t\) maxima provide simultaneous bounds over languages for H1 and over budgets for H3; H2 uses the corresponding one-sided studentized bootstrap contrast. Because calibration with 250 discrete item clusters found mild anti-conservatism in the extreme \(\alpha/6\) tail, every raw confirmatory p-value is multiplied by the pre-specified 1.3 tail-conservatism factor (capped at 1), equivalently testing at \(\alpha/1.3\), before Holm correction.
The corrected type-I rate was 0.00917 against a 0.00833 target; this exceeds the target and is consistent with nominal only within Monte-Carlo tolerance (SE 0.0012), so the factor is a conservative safeguard rather than a verified family-wise calibration.

The independent-decoding replication has its own protocol and its own family of six tests, specified in \texttt{prereg-independent-decoding.md} at tag \texttt{independent-protocol-freeze}, with results in \texttt{analysis-out/independent\_scoring.json}. It reuses this machinery unchanged, except that \(\operatorname{acc}_N(B)\) and \(\operatorname{acc}_N(\lfloor rB\rfloor)\) now come from different generations and so are paired within item rather than within trace.

\section{TRANSLATE-ACT translation quality (COMET, descriptive)}

Reference-based \texttt{Unbabel/wmt22-comet-da}; one stored full trace per item (sample 0); delimiter-missing traces excluded from this denominator only; pointwise 95\% item-bootstrap CIs (10,000 resamples).

\begin{table*}[t]
\centering
\small
\caption{Reference-based COMET translation quality for TRANSLATE-ACT. Intervals are pointwise 95\% item-bootstrap CIs.}
\label{tab:comet}
\resizebox{\textwidth}{!}{%
\begin{tabular}{llrrrrrrr}
\toprule
Model & Lang & n & Missing delimiter & COMET mean & median & p10 & p90 & 95\% CI \\
\midrule
Qwen  & de & 250 & 0.0\% & 0.877 & 0.878 & 0.834 & 0.921 & [0.872, 0.881] \\
Qwen  & th & 249 & 0.4\% & 0.858 & 0.866 & 0.810 & 0.907 & [0.851, 0.865] \\
Qwen  & sw & 250 & 0.0\% & 0.749 & 0.772 & 0.580 & 0.867 & [0.734, 0.763] \\
Llama & de & 249 & 0.4\% & 0.872 & 0.874 & 0.826 & 0.915 & [0.868, 0.877] \\
Llama & th & 244 & 2.4\% & 0.783 & 0.850 & 0.325 & 0.895 & [0.759, 0.805] \\
Llama & sw & 240 & 4.0\% & 0.798 & 0.825 & 0.703 & 0.888 & [0.783, 0.811] \\
\bottomrule
\end{tabular}
}
\end{table*}

These scores are descriptive only and never condition, gate, or reweight accuracy.

\begin{table}[t]
\centering
\small
\caption{Exploratory COMET association with correctness gain at each prespecified peak token budget, using sample 0. Intervals are pointwise bootstrap 95\% CIs.}
\label{tab:comet-gain}
\resizebox{\columnwidth}{!}{%
\begin{tabular}{llrrr}
\toprule
Model & Lang & Peak \(B\) & \(n\) & Spearman \(\rho\) [95\% CI] \\
\midrule
Qwen  & de & 192 & 250 & 0.015 [\(-0.112\), 0.139] \\
Qwen  & th & 256 & 249 & 0.146 [0.019, 0.269] \\
Qwen  & sw & 128 & 250 & \(-0.143\) [\(-0.269\), \(-0.004\)] \\
Llama & de & 256 & 249 & 0.280 [0.155, 0.399] \\
Llama & th & 192 & 244 & 0.009 [\(-0.129\), 0.146] \\
Llama & sw & 256 & 240 & 0.074 [\(-0.051\), 0.203] \\
\bottomrule
\end{tabular}
}
\end{table}

\section{Vocabulary-extension measurement}
\label{app:vocab}

The extension procedure is described in \S\ref{sec:ladder}. Cross-fitting runs
over two disjoint halves of the 250 MGSM items, so no evaluated item
contributed a merge to the tokenizer that scores it, for either arm. The base
pipeline reproduces the frozen premiums exactly (1.5589 / 2.5508 / 1.9363).

The English control compares aggregate English devtest token counts before and
after extension. It bounds the aggregate cost below \(0.02\%\), but appending
merges guarantees only that base merges retain priority; it does not guarantee
that every English string segments identically, and a small number do change.

\begin{table}[t]
\centering
\small
\caption{Qwen vocabulary extension, per cross-fitting fold. \(r\) is the frozen
FLORES-200 premium and \(r'\) is the premium under the extended tokenizer. The
English control is the aggregate English devtest token count relative to the
base tokenizer. These are the extensions used in Table~\ref{tab:ladder}.}
\label{tab:vocab}
\begin{tabular}{llrrrr}
\toprule
Lang & Fold & New tokens & \(r\) & \(r'\) & English \\
\midrule
de & 0 & 3{,}470 & 1.559 & 1.531 & 0.99996 \\
de & 1 & 3{,}343 & 1.559 & 1.532 & 0.99996 \\
th & 0 & 6{,}798 & 2.551 & 2.195 & 1.00000 \\
th & 1 & 6{,}577 & 2.551 & 2.218 & 1.00000 \\
sw & 0 & 3{,}173 & 1.936 & 1.865 & 0.99993 \\
sw & 1 & 3{,}054 & 1.936 & 1.826 & 0.99986 \\
\bottomrule
\end{tabular}
\end{table}

Extension sizes are limited by our corpora, which are MGSM reasoning traces
rather than general target-language text; a production-scale extension trained
on a broader corpus would plausibly compress more, but we cannot bound that
from these data, so we report the compression observed under this corpus rather
than a lower bound on what is achievable.

Training text is the NATIVE arm only: the PIVOT and CODE-SWITCHED arms are
substantially English in several cells (Swahili PIVOT is 90.1\% English,
\S\ref{sec:audits}) and would not train a language-specific vocabulary. Because
TRANSLATE-ACT traces are largely English, a target-language extension barely
compresses them, so their gains are correspondingly small; that is a property
of the intervention rather than an artifact of the design. The counterfactual
holds the emitted text fixed and changes only what that text costs. A real
extension must also learn embeddings for the new tokens
\citep{chinesellama,swallow,focus}, which is itself a training procedure, and a
model whose tokenizer changed would follow a different trajectory.

The Swahili rows of Table~\ref{tab:ladder} show why the truncated share alone
is misleading. At \(B=1024\), 11.6\% of NATIVE traces exceed the cap, but 211
of those 232 traces run to the 4096-token generation cap without stopping, only
23 are parseable even from the full stored trace, and only two become correct
by 4096. The NATIVE gain to 4096 is therefore 0.10 points, of which the
extension recovers 0.05. That cell is not evidence that extension is weak
against real truncation; it is a cell with almost nothing left to recover
within the ledger.

Two cautions constrain how far \S\ref{sec:ladder} generalizes. Our ledger caps
generation at 4096 tokens, so \(G(4096)\) is the gap at the largest prefix we
stored, not a demonstrated non-binding regime; 10.55\% of Swahili NATIVE
generations reach that cap without stopping. The scale of the intervention is
also modest by construction: cutting the Thai premium from 2.551 to 2.207 is
13.5\% fewer tokens for the same text, equivalently 15.6\% more text under a
fixed cap, against the four- to thirty-two-fold increases available by raising
the cap itself. Vocabulary extension may also reduce decoding steps, though we
did not measure net serving cost or latency, and a larger embedding and output
projection work against the saving. The fold split is fixed rather than
randomized and we did not run repeated splits, and the emission percentiles
in Table~\ref{tab:emission-timing} are computed among emitting traces only, covering
neither the 1.95--25.1\% that never emit nor the upper tenth that emit late.

Accuracy in Table~\ref{tab:ladder} does not use \(\rho\) as a multiplier; the
retokenization path is the one described in \S\ref{sec:ladder}.
Decoding token ids, rather than slicing the source string at a token offset,
matters because byte-level tokens can split a code point, so an offset slice
would occasionally admit slightly more than \(B\) tokens. Scoring generated
text rather than stored token ids is a reimplementation of the scorer used
elsewhere in the paper, so we verified it against that scorer on all six Qwen
cells at five budgets. The two paths agree on 59{,}998 of 60{,}000
sample-budget comparisons, the two exceptions being Swahili NATIVE traces whose
text does not round-trip to its stored token ids. Every column of
Table~\ref{tab:ladder} is produced by the retokenization path alone, so the
reported closure is exactly \(G-G_3\). No model weights are trained.

\section{Budget announcement (E2)}
\label{app:announce}

The E2 ledger contains 876,000 records over 438 shards, spanning both a coupled block, in which the
announced number equals the enforced cap across the budget grid, and a fixed-cap block. The confirmatory
contrast uses only the fixed-cap block: 192,000 records over 96 shards, all enforced at 2048, in which
only the announced number varies over \(\{128,256,2048\}\), so truncation is held constant across the
contrast.
The estimand is
\(\Delta_{\mathrm{ann}} = \operatorname{acc}^{128}(2048)-\operatorname{acc}^{2048}(2048)\),
two-sided, Holm step-down at family-wise \(\alpha=0.05\) with first-step local
\(\alpha_1=0.0125\). Every raw p-value carries the same \(1.3\times\)
tail-conservatism factor used elsewhere in the paper. The announced-256 cell is
an interpolation and sits outside the family by design.

\begin{table}[t]
\centering
\small
\caption{The E2 confirmatory family: four two-sided announcement dose contrasts
on Qwen3-8B at a fixed enforced cap of 2048. Formal outcome
\texttt{announcement\_effect\_detected}.}
\label{tab:e2}
\resizebox{\columnwidth}{!}{%
\begin{tabular}{lllrrrrrl}
\toprule
test & arm & lang & acc @128 & acc @2048 & \(\Delta_{\mathrm{ann}}\) & SE & \(p\) & reject \\
\midrule
A1-nat-de & NATIVE & de & 78.05 & 80.90 & \(-2.85\) & 1.29 & 0.0380 & no \\
A1-nat-th & NATIVE & th & 63.20 & 58.10 & \(+5.10\) & 1.67 & 0.0029 & \textbf{yes} \\
A1-ta-de  & TR-ACT & de & 87.15 & 87.80 & \(-0.65\) & 0.69 & 0.4776 & no \\
A1-ta-th  & TR-ACT & th & 87.70 & 86.25 & \(+1.45\) & 0.75 & 0.0747 & no \\
\bottomrule
\end{tabular}
}
\end{table}

\begin{table}[t]
\centering
\small
\caption{Exploratory dose response over the announced grid under the
announcement instrument, Qwen3-8B. Median output tokens accompany each
accuracy. Thai NATIVE is monotone in the announced number.}
\label{tab:e2-dose}
\resizebox{\columnwidth}{!}{%
\begin{tabular}{llrrr}
\toprule
arm & lang & announced & accuracy & median tokens \\
\midrule
NATIVE & de & 128  & 78.05 & 177 \\
NATIVE & de & 256  & 77.20 & 214 \\
NATIVE & de & 2048 & 80.90 & 292 \\
\midrule
NATIVE & th & 128  & 63.20 & 198 \\
NATIVE & th & 256  & 59.90 & 239 \\
NATIVE & th & 2048 & 58.10 & 350 \\
\midrule
TR-ACT & de & 128  & 87.15 & 222 \\
TR-ACT & de & 256  & 87.15 & 233 \\
TR-ACT & de & 2048 & 87.80 & 260 \\
\midrule
TR-ACT & th & 128  & 87.70 & 264 \\
TR-ACT & th & 256  & 87.80 & 269 \\
TR-ACT & th & 2048 & 86.25 & 293 \\
\bottomrule
\end{tabular}
}
\end{table}

The machine-readable tag arm announces the same numbers as
\texttt{TOKEN\_BUDGET: \{budget\}} and changes median length by at most 2.8\% in absolute value in the
twelve cells it covers, rejecting in none of them.

Budget forcing injects the answer delimiter when the cap arrives. Its two
populations must be read separately: \texttt{capped\_eos} false is a trace the
cap truncated, and \texttt{capped\_eos} true is a trace that completed and
still emitted no answer line, where forcing repairs a formatting failure rather
than relieving a budget. For Qwen NATIVE German at \(B=128\), 97.00\% of traces
are forced, 99.74\% of those by truncation; pooled forced accuracy is 25.70\%
against the 2.55\% blind (no-announcement) baseline, decomposing into 23.72\%
among truncated traces and
100.00\% among the completed-but-unformatted ones. These are exploratory.

Absolute levels in this family are not comparable to the sweep ledger: every
confirmatory cell is freshly sampled at an enforced 2048 cap and carries the
announcement instrument in its prompt, so Thai NATIVE at announced 2048
(58.10\%) sits above the sweep plateau at \(B=1024\) (47.1\%). The
confirmatory contrast is within-instrument and unaffected.

\section{Instrument validity (E2b)}
\label{app:instrument}

The family, estimand, announced values, enforced cap, and Holm structure are
unchanged from Appendix~\ref{app:announce}; only the TRANSLATE-ACT instruction
changed. Instrument v0 reads ``The translation, all of your reasoning and the
final answer may take at most \{budget\} tokens in total.'' Instrument v1 reads
``Your entire response must not exceed \{budget\} tokens. Keep the translation
as short as possible, reason concisely, and write the \#\#\#\# line before you
reach the limit.'' NATIVE is unchanged and its records are the same under both
rows, so the NATIVE rows are one measurement printed twice rather than a
replication.

\begin{table}[t]
\centering
\small
\caption{The same four cells under both TRANSLATE-ACT instruments, Qwen3-8B.
The gate is a 30\% median-length reduction between announced budgets; a cell
that fails it is uninformative about budget sensitivity and is not evidence of
no effect. No Holm decision differs between the instruments and the formal
outcome is unchanged.}
\label{tab:e2b}
\resizebox{\columnwidth}{!}{%
\begin{tabular}{lllrrrl}
\toprule
test & inst. & \(\Delta_{\mathrm{ann}}\) & SE & \(p\) & median red. & reading \\
\midrule
A1-nat-de & v0 & \(-2.85\) & 1.29 & 0.0380 & 39.4\% & interpretable \\
A1-nat-de & v1 & \(-2.85\) & 1.29 & 0.0380 & 39.4\% & interpretable \\
A1-nat-th & v0 & \(+5.10\) & 1.67 & 0.0029 & 43.4\% & interpretable \\
A1-nat-th & v1 & \(+5.10\) & 1.67 & 0.0029 & 43.4\% & interpretable \\
A1-ta-de  & v0 & \(-0.65\) & 0.69 & 0.4776 & 14.6\% & uninformative \\
A1-ta-de  & v1 & \(-2.60\) & 1.09 & 0.0229 & 34.4\% & interpretable \\
A1-ta-th  & v0 & \(+1.45\) & 0.75 & 0.0747 & 10.1\% & uninformative \\
A1-ta-th  & v1 & \(-2.30\) & 1.17 & 0.0662 & 37.5\% & interpretable \\
\bottomrule
\end{tabular}
}
\end{table}

In the E2b pilot, the translation segment remains at 57 tokens in German and
76 in Thai across announced values under v0, a 0.0\% response in both
languages, and changes by 5.3\% and 16.0\% under v1; the pilot's total median
reductions are 34.1\%/36.8\% against 14.6\%/9.9\%. The study-ledger totals are
those of Table~\ref{tab:e2b}: 34.4\%/37.5\% under v1 and 14.6\%/10.1\% under
v0.

Llama-3.1-8B-Instruct fails the 30\% gate in all four cells under both
instruments, with median reductions of 2.4--9.3\% against Qwen's 34--43\%.
Every Llama estimate in this family is therefore uninformative about budget
sensitivity, including its two rejections (NATIVE de \(-4.35\), NATIVE th
\(+2.50\)), which we do not read as budget findings. Llama carries no
confirmatory claim here.

\section{Correct-emission sub-CDF consistency check}
\label{app:subcdf}

For NATIVE, define the correct-emission sub-CDF
\(G(t)=P(C=1,E\le t)\), where \(E\) is answer-emission position and \(C\) is
correctness of the completed trace. Under Equation~(1),
\[
\Delta_L(B)=G(\lfloor rB\rfloor)-G(B).
\]
Both terms are estimated from the same stored 4096-cap ledger. The peak
budgets were selected from the discovery sweep, so this three-cell comparison
is exploratory. It is a consistency check, not a test.

\begin{table}[t]
\centering
\small
\caption{Correct-emission sub-CDF predictions from the Qwen3-8B NATIVE MGSM
ledger and observed deltas from the separately generated independent sweep.
Values are percentage points.}
\label{tab:subcdf}
\begin{tabular}{llrrr}
\toprule
lang & window & observed & predicted & error \\
\midrule
de & \((192,299]\) & 34.65 & 34.20 & \(-0.45\) \\
th & \((256,652]\) & 38.60 & 38.85 & \(+0.25\) \\
sw & \((128,247]\) & 13.70 & 14.95 & \(+1.25\) \\
\bottomrule
\end{tabular}
\end{table}

The mean absolute error against independent-decoding outcomes is 0.65 points. The independent-decoding standard errors are 2.10 / 2.26 / 1.37, making
the absolute residuals 0.21 / 0.11 / 0.91 outcome SEs. Agreement with REPLAY
deltas is not evidence: under absorbing correctness, the sub-CDF on a replay
ledger is algebraically identical to the replay accuracy difference and
therefore agrees by construction.

An exploratory replay-frame breadth analysis applies the same construction to
Qwen3-8B NATIVE long-cap ledgers on three further benchmarks. Within each cell,
\(G\) is estimated on even-indexed items and \(\Delta_L(B)\) is scored on
odd-indexed items, making the halves disjoint. Same-item scoring would be
circular: under absorbing correctness,
\(G(\lfloor rB\rfloor)-G(B)\) is the prefix-scored accuracy difference itself,
so agreement would be guaranteed.

\begin{table*}[t]
\centering
\small
\caption{Exploratory split-half correct-emission sub-CDF results from Qwen3-8B
NATIVE long-cap ledgers. MAE and peaks are percentage points; peak suffixes
give the token budget.}
\label{tab:subcdf-breadth}
\begin{tabular}{llrrrr}
\toprule
benchmark & lang & \(r\) & MAE & observed peak & predicted peak \\
\midrule
MMATH & es & 1.522 & 1.77 & 20.65 @384 & 19.03 @256 \\
MMATH & th & 2.551 & 1.15 & 36.45 @384 & 35.95 @384 \\
Belebele & de & 1.559 & 0.67 & 40.31 @256 & 36.64 @256 \\
Belebele & th & 2.551 & 0.37 & 21.11 @384 & 21.39 @384 \\
Belebele & sw & 1.936 & 0.44 & 10.83 @64 & 12.53 @64 \\
Global-MMLU-Lite & de & 1.559 & 0.81 & 32.94 @384 & 31.81 @384 \\
Global-MMLU-Lite & sw & 1.936 & 1.22 & 7.50 @64 & 5.94 @128 \\
\bottomrule
\end{tabular}
\end{table*}

Across these seven cells, the mean absolute error is 0.92 points on held-out
items. Peak location is exact in five of seven; the other two differ by one grid
point. This design differs from MGSM: there the predictor comes from the replay
ledger and the outcome from a separately generated independent sweep, whereas
here both halves come from one ledger. The 0.92 and 0.65 errors are therefore
not like-for-like and do not establish a trend.

MMATH zh is separate and is not counted among the seven. Its premium is
\(r=1.003\), so \((B,\lfloor rB\rfloor]\) is about one token wide and
\(\Delta_L(B)\) is approximately zero throughout by construction (observed
peak 0.14; MAE 0.05). This is a structural check that the estimand vanishes
without a premium, not a successful prediction of the mechanism.

These results are Qwen-only and exploratory, use item-level split halves, and
remain in the replay frame. Llama parse rates on these benchmarks ranged from
0.1\% to 29\%, producing one usable cell of eight. Its prose answers lack the
required \texttt{\#\#\#\#} form, and the required log-probability data were
unavailable, so Llama could not be scored on these added benchmarks.

For MGSM, the naive factorization
\(p_{\mathrm{correct}}\times[F_E(\lfloor rB\rfloor)-F_E(B)]\) predicts
30.41 / 36.69 / 10.53, for a mean absolute error of 3.10 points. It
incorrectly assumes that correctness and emission time are independent. Of 6,000 records, 697 never emit a parseable answer at this analysis's probe resolution; these non-emitters are 0\% correct by construction, against 60.3\% correctness among emitters.

In MGSM, correctness is not strictly absorbing because \texttt{parse\_answer} reads the
last answer line. On the same ledger, genuine answer revision occurs in
1.35\% of records and correct-to-wrong revision in 0.52\%. Of apparent
instability, 98\% is the parser reading a number mid-write; this cannot bias
\(\Delta_L(B)\) because such a prefix scores wrong under both frames.

The MGSM calculation covers only three Qwen3-8B cells and supports no claim
across models or benchmarks.

\end{document}